# Singularity Surfaces and Maximal Singularity-Free Boxes in the Joint Space of Planar 3-R<u>P</u>R Parallel Manipulators

Mazen ZEIN          Philippe WENGER          Damien CHABLAT
IRCCyN, Institut de recherche en communications et cybernéthique de Nantes
1, rue de la Noe, 44322 Nantes

**Abstract**— *In this paper, a method to compute joint space singularity surfaces of 3-RPR planar parallel manipulators is first presented. Then, a procedure to determine maximal joint space singularity-free boxes is introduced. Numerical examples are given in order to illustrate graphically the results. This study is of high interest for planning trajectories in the joint space of 3-RPR parallel manipulators and for manipulators design as it may constitute a tool for choosing appropriate joint limits and thus for sizing the link lengths of the manipulator.*

Keywords: *3-R<u>P</u>R* parallel manipulators, singularity, singularity-free zones, joint space, joint limits.

## I. Introduction

Most parallel manipulators have singularities that limit the motion of the moving platform. The most dangerous ones are the singularities associated with the direct kinematics, where two assembly-modes coalesce. Indeed, approaching such a singularity results in large actuator torques or forces, and in a loss of stiffness. Hence, these singularities are undesirable. There exists three main ways of coping with singularities, which have their own merits. A first approach consists in eliminating the singularities at the design stage by properly determining the kinematic architecture, the geometric parameters and the joint limits [4,4,7]. This approach is safe but difficult to apply in general and restricts the design possibilities. A second approach is the determination of the singularity-free regions in the workspace [5,15-17,20,24]. This solution does not involve a priori design restrictions but, because of the complexity of the singularity surfaces, it may be difficult to determine definitely safe regions. Finally, a third way consists in planning singularity-free trajectories in the manipulator workspace [2,6,19]. With this solution one is also faced with the complexity of the singularity equations but larger zones of the workspace may be exploited.
In this paper, we choose to use the second approach by determining maximal joint space singularity-free boxes. This approach will help us determine appropriate joint limits and link dimensions.
Planar parallel manipulators and particularly manipulators with three extensible leg rods, referred to as 3-R<u>P</u>R, have received a lot of attention because they have interesting potential applications in planar motion systems [9,21]. As shown in [18], moreover, the study of the 3-R<u>P</u>R planar manipulator may help better understand the kinematic behavior of its more complex spatial counterpart, the 6-dof octahedral manipulator, which has also triangular base and platforms.
The singularities of these manipulators have been most often represented in their workspace [13,14,18] but more rarely in their joint space [18,24,26].
Hunt and Primrose showed that 3-R<u>P</u>R planar manipulator could have up to 6 assembly-modes [12]. Mcaree and Daniel analyzed the joint space singularities through slices to explain non-singular changing trajectories [188], and Zein et al analyzed the topology of these slices in [26]. It was shown in [18,24,25] that, to change its assembly mode without meeting a singularity, a *3-R<u>P</u>R* manipulator should encircle a cusp point in its joint space.
In this paper, a method to compute and to represent joint space singularities of 3-R<u>P</u>R planar parallel manipulators is first proposed. A procedure is then provided to determine maximal joint space singularity-free boxes.
This work is of a high interest for the determination of appropriate joint limits and for planning trajectories in the joint space.

## II. Manipulators under study

The manipulators under study are 3-DOF planar parallel manipulators with three extensible leg rods (Fig.1). These manipulators have been frequently studied and have interesting potential applications in planar motion systems. The geometric parameters are the three sides of the moving platform $d_1$, $d_2$, $d_3$ and the position of the base revolute joint centers defined by $A_1$, $A_2$ and $A_3$. The reference frame is centered at $A_1$ and the *x*-axis passes through $A_2$. Thus, $A_1 = (0, 0)$, $A_2 = (A_{2x}, 0)$ and $A_3 = (A_{3x}, A_{3y})$. The parameter $\beta$ is function of $d_1$, $d_2$ and $d_3$.
The joint space $Q$ is defined by the vectors of the lengths of the three actuated extensible links $\dot{\mathbf{q}} = \begin{bmatrix} \dot{\rho}_1 & \dot{\rho}_2 & \dot{\rho}_3 \end{bmatrix}^T$.

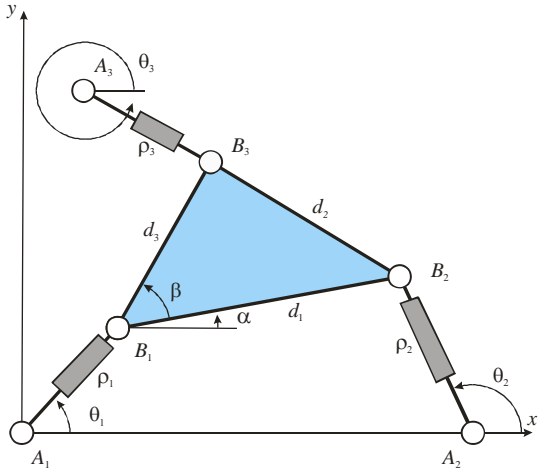

Fig. 1.  A 3-R<u>P</u>R parallel manipulator

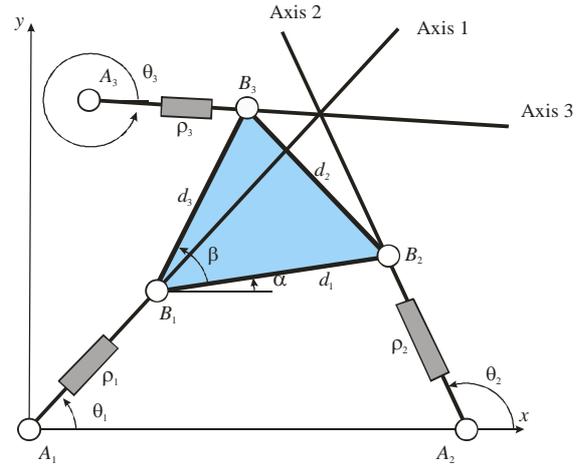

Fig. 2.  A 3-R<u>P</u>R parallel manipulator on a singular configuration.

### III. Kinematics of 3-R<u>P</u>R parallel manipulators

The relation between the joint space **Q** and the workspace **W** can be expressed as a system of non-linear algebraic equations, which can be written as:

$$F(\mathbf{x}, \mathbf{q}) = 0 \quad (1)$$

where **x** and **q** are respectively the vectors of the workspace and joint space variables.

Differentiating equation (1) with respect to time leads to the velocity model:

$$\mathbf{A t} + \mathbf{B \dot q} = \mathbf{0} \quad (2)$$

where $\mathbf{t} = [w \; \dot{\mathbf{c}}]^T$ (w is the scalar angular velocity and $\dot{\mathbf{c}}$ is the two-dimensional velocity vector of the operational point $B_1$ of the platform if we used the first workspace parameters), **A** and **B** are 3×3 Jacobian matrices which are configuration dependent, and $\dot{\mathbf{q}} = [\dot\rho_1 \; \dot\rho_2 \; \dot\rho_3]^T$ is the joint velocity vector.

### IV. Joint space singularities of 3-R<u>P</u>R parallel manipulators

The singularities of 3-DOF planar parallel manipulators have been extensively studied (see for example [3,8,14,18,22]). They were defined in the workspace (*x, y, α*) and to the author's knowledge, there exist a small number of works dealing with the singular configurations in the manipulators joint space ($\rho_1, \rho_2, \rho_3$).

In a parallel singularity, matrix **A** is singular. To derive the singularity equations, it is usual to expand the determinant of **A**. We use rather a geometric approach that does not involve complicated algebraic calculus. The *3-R<u>P</u>R* parallel manipulator is in a singular configuration whenever the axes of its three legs are concurrent or parallel [11] (Fig. 2).

In order to derive this geometric condition, we derive the condition for the three leg axes to intersect at a common point (possibly at infinity). We first write the equations of the three leg axes:

$$\begin{cases} (\text{Axis 1}): y\cos(\theta_1) = x\sin(\theta_1) \\ (\text{Axis 2}): y\cos(\theta_2) = (x - A_{2x})\sin(\theta_2) \\ (\text{Axis 3}): y\cos(\theta_3) = (x - A_{3x})\sin(\theta_3) + A_{3y}\cos(\theta_3) \end{cases} \quad (3)$$

Eliminating *x* and *y* yields the following singularity equation in the task parameters ($\theta_1, \theta_2, \theta_3$):

$$A_{2x}s_2 s_{31} + (A_{3x}s_3 - A_{3y}c_3)s_{12} = 0 \quad (4)$$

where $s_i = \sin(\theta_i)$, $c_i = \cos(\theta_i)$ and $s_{ij} = \sin(\theta_i - \theta_j)$.

It is possible to express Eq. (4) as a function of the joint space parameters $\rho_1$, $\rho_2$ and $\rho_3$ by using the constraint equations of the *3-R<u>P</u>R* manipulator. However, the resulting equation would be too complicated to yield real insights, and difficult to handle.

Our approach to compute the singular configurations in the joint space consists in reducing the dimension of the problem by first considering two-dimensional slices of the configuration space by fixing the first leg rod length $\rho_1$. The singular surfaces in the full joint space are then calculated by "stacking" the slices.

***Step 1:*** We rewrite Eq. (4) as a function of $\rho_1$, $\alpha$ and $\theta_1$ using the constraint equations of the manipulator.

$$\begin{cases} A_{2x} + \rho_2 c_2 - \rho_1 c_1 - d_1 \cos(\alpha) = 0 \\ \rho_2 s_2 - \rho_1 s_1 - d_1 \sin(\alpha) = 0 \\ A_{3x} + \rho_3 c_3 - \rho_1 c_1 - d_3 \cos(\alpha + \beta) = 0 \\ A_{3y} + \rho_3 s_3 - \rho_1 s_1 - d_3 \sin(\alpha + \beta) = 0 \end{cases} \quad (5)$$

The first (respectively last) two equations make it possible to express $\rho_2$ (respectively $\rho_3$) as function of $\rho_1$, $\alpha$ and $\theta_1$. Then, $c_2$ and $s_2$ (respectively $c_3$ and $s_3$) are calculated as function of $\rho_1$, $\alpha$ and $\theta_1$ from the first (respectively last) two equations of (5) and their expressions are input in Eq. (4), which, now, depend only on $L_1$, $\alpha$ and $\theta_1$.

***Step 2:*** We fix a value for $\rho_1$, so Eq. (4) depends now only on $\alpha$ and $\theta_1$. By varying $\alpha$ or $\theta_1$, we compute the roots of the equation, to obtain the singular configurations ($\alpha_s$, $\theta_{1s}$) for a fixed $\rho_{1s}$.

***Step 3:*** For every singular configuration computed in the space ($\alpha$, $\theta_1$) in the second step of the approach, we calculate the corresponding values $\rho_{2s}$ and $\rho_{3s}$ using the equation system (6). We have thus the singular configurations curves in a slice of the joint space ($\rho_2$, $\rho_3$) with $\rho_1$ fixed.

$$\begin{cases} \rho_2 = \sqrt{\left(-A_{2x} + \rho_1 \cos(\theta_1) + d_1 \cos(\alpha)\right)^2 + \left(\rho_1 \sin(\theta_1) + d_1 \sin(\alpha)\right)^2} \\ \rho_3 = \sqrt{\left(A_{3x} - \rho_1 \cos(\theta_1) - d_3 \cos(\alpha+\beta)\right)^2 + \left(A_{3y} - \rho_1 \sin(\theta_1) - d_3 \sin(\alpha+\beta)\right)^2} \end{cases} \quad (6)$$

Figure 3 shows a slice of the joint space singular configurations for $\rho_1$=17 obtained for the same 3-*RPR* manipulator used in [14,18,21]. We refer only to this manipulator in this paper in order to illustrate our work. The geometric parameters of this manipulator are recalled below in an arbitrary length unit:

| $A_1$=(0, 0) | $d_1$=17.04 |
|---|---|
| $A_2$=(15.91, 0) | $d_2$=16.54 |
| $A_3$=(0, 10) | $d_3$=20.84 |

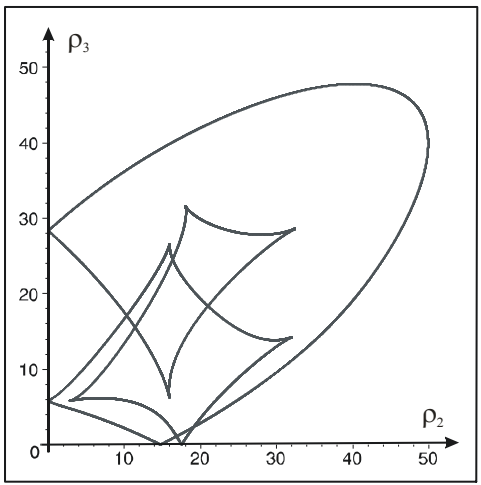

Fig. 3. Singular configurations in ($\rho_2$, $\rho_3$) for $\rho_1$=17.

***Step 4:*** We compute the joint space singularity slices for a number of $\rho_1$ values, to do this we have to repeat steps 2 and 3 while varying $\rho_1$.

Finally, we collect all the computed slices in one file to obtain the singularities in the joint space ($\rho_1$,$\rho_2$, $\rho_3$).

Figure 4 represents the singularities in the joint space of the manipulator studied when $\rho_1$ varies from 0 to 50. To obtain this surface, we have imported the solutions obtained in step 4 into a CAD software, and we have meshed them together.

Obviously, there is continuity between the singularities slices, one can claim, without any doubt, that there is singularity between the different slices.

The surface depicted in Fig. 4 is of interest:

***i.*** for planning trajectories in the joint space because it shows clearly the joint space regions that are free of singularities.

***ii.*** for manipulator design, because it constitutes a tool for defining the values of the joint limits such that the joint space is a singularity-free box.

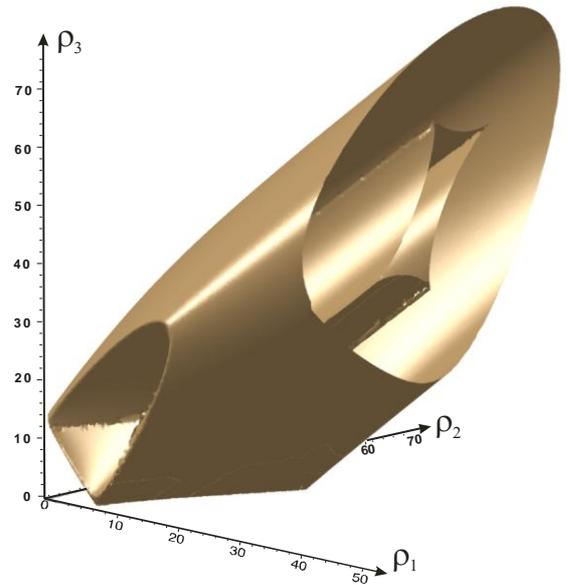

Fig. 4. Joint space singularity surfaces of the 3-*RPR* manipulator studied when $\rho_1$ varies from 0 to 50.

## V. Maximal joint space singularity-free boxes

In a context of design and/or trajectory planning, an important problem is to find singularity-free zones in the joint space.

In this section, we introduce a new procedure to determine maximal singularity-free boxes in the joint space of *3-RPR* manipulators. These singularity-free boxes will help us fix the manipulator joints limits.

Two numerical examples are provided to illustrate the effectiveness of the procedure.

### A. Procedure

***Step 1:*** We choose an initial joint space configuration $Q_0(\rho_{10},\rho_{20},\rho_{30})$. This configuration can be chosen according to several considerations, for example choosing $Q_0$ as the image through the inverse kinematics of a prescribed workspace center, or choosing it directly in the joint space as the center of a large singularity-free zone.

***Step 2:*** We calculate the largest singularity-free cube centered at $Q_0(\rho_{10},\rho_{20},\rho_{30})$.

To do this, we calculate the infinity norm distance $d$, also known as Chebyshev distance, between the center point $Q_0(\rho_{10},\rho_{20},\rho_{30})$ and each of the joint space singular points $Q_s(\rho_{1s},\rho_{2s},\rho_{3s})$ computed in Section IV:

$$d = \max\left(|\rho_{10}-\rho_{1s}|,|\rho_{20}-\rho_{2s}|,|\rho_{30}-\rho_{3s}|\right) \qquad (7)$$

and we keep the minimal distance $d_{min}$ found over all, because we are searching for the distance between the closest joint space singularity configuration $Q_s$ from the center point $Q_0$.

The length of the singularity-free cube edge $a$ will be:

$$a = 2 \times d_{min} \qquad (8)$$

***Step 3:*** The choice of the initial center point $Q_0(\rho_{10},\rho_{20},\rho_{30})$ does not lead to an optimized solution, in other words varying lightly the center point position may lead to a largest singularity-free cube. Thus, the position of the initial point must be optimized, which we have done using a Hooke and Jeeves optimization scheme [10]. Note that the solution found is a local optimum.

***Step 4:*** The cube found in step 3 touches the closest singular configuration to the center point. In order that the cube does not touch the singularities surface and for more security we subtract a small security value $s$ from the distance $d_{min}$. Such a value can be related to laws of command to stop the motion when the joint velocity is maximum.

The manipulator joint limits corresponding to the cube found can be easily computed as follows:

$$\begin{cases}\rho_{i\min} = \rho_{i0}-(d_{min}-s)\\ \rho_{i\max} = \rho_{i0}+(d_{min}-s)\end{cases} \text{ with } i=1,2,3 \qquad (9)$$

### B. Application of the procedure

In this section, two examples are provided in order illustrate the application of the procedure.

***Example 1:***
For the same manipulator studied, we consider the center point $Q_0(35,25,45)$ in the joint space. This point was chosen in the center of a large singularity-free zone in the joint space. By computing the Chebyshev distances between each joint space singular point $Q_s$ computed in section IV, and $Q_0$, the minimal distance obtained is $d_{min}=5.3$, so the edge length of the singularity-free cube is $a = 10.6$.
By running the optimization algorithm, we find a maximal value $d_{min}=7.175$ for a center point $Q(41.625, 24.875, 44.125)$.
We subtract a security value of 0.1 from $d_{min}$ which becomes $d_{min}=7.075$.
Figure 5 shows the joint space singularity surface and the maximal joint space singularity-free cube centered at $Q(41.625, 24.875, 44.125)$ for the manipulator studied.

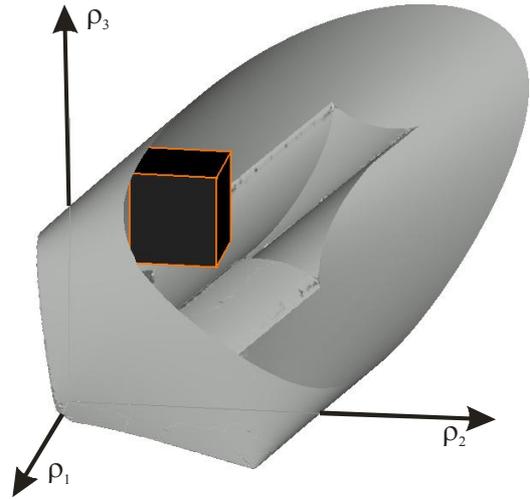

Fig. 5. Joint space singularity surfaces and maximal joint space singularity-free cube centered at Q(*41.625, 24.875, 44.125*).

Figures 6 shows the images through the direct kinematics of the maximal joint space singularity-free cube, which are two separate singularity-free components, each of them being located in an aspect of the workspace [24]. The projections of these two components onto the (x,y) plane are plotted in gray in Figure 6.

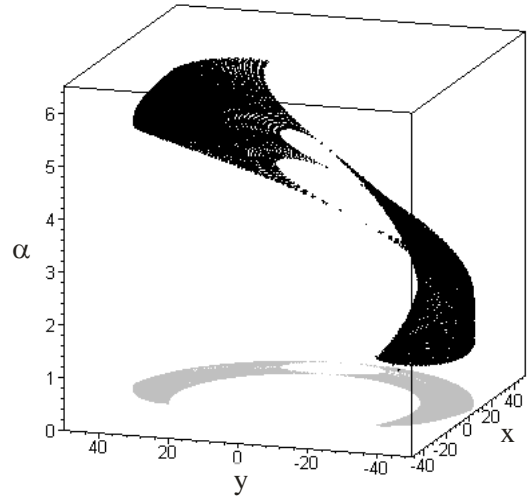

Fig. 6. Images by direct kinematics of the joint space singularity-free cube (in black), and their projection on the (x,y) plane (in gray).

Figures 7 shows the two workspace components and the workspace singularities of the *3-RPR* manipulator studied, the singularities are plotted in color.

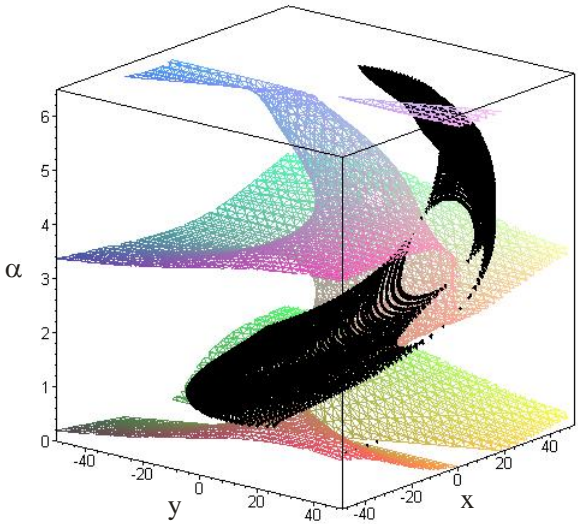

Fig. 7. Singularity-free components with workspace singularities.

*Example 2:*

For the same manipulator, we consider in this example the center point $Q_0(30,50,35)$ in the joint space. By computing the Chebyshev distances between each joint space singular point $Q_s$ computed in Section IV and $Q_0$, the minimal distance obtained is $d_{min}=4$, so the edge length of the singularity-free cube is $a = 8$.

By running the optimization algorithm, we find a maximal value $d_{min}=5.794$ for a center point $Q(38.125, 50, 33)$.

We subtract a security value of 0.1 from $d_{min}$, which becomes $d_{min}=5.694$.

Figure 8 shows the joint space singularity surface and the maximal joint space singularity-free cube centered at $Q(38.125, 50, 33)$ for the manipulator studied.

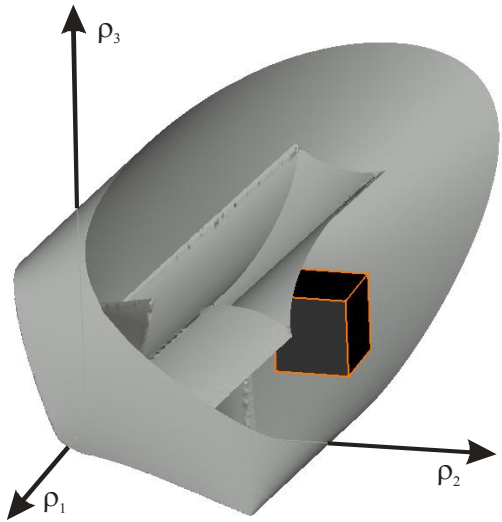

Fig. 8. Joint space singularity curves and maximal joint space singularity-free cube centered at $Q(38.125, 50, 33)$.

Figures 9 displays the images by the direct kinematics of the maximal joint space singularity-free cube, which are two separate singularity-free components, one in each aspect of the workspace. The projections of these two components onto the (x,y) plane are plotted in gray in Figure 9. This figure is displayed with the same viewing angle as in Figure 6 to show the difference between the two examples.

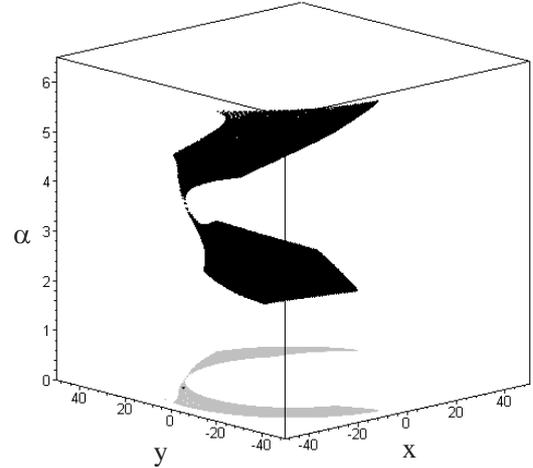

Fig. 9. Images by direct kinematics of the joint space singularity-free cube (in black), and their projection on the (x,y) plane (in gray).

### C. Future works

We can see in figures 6 and 9 that the components in the workspace do not have regular forms. Because this study is carried out in the joint space only, it cannot take into account any of the properties, in the workspace, of the image by direct kinematics of the singularity-free cube found.

This work will be extended by taking into account the largest regular volume (cube, cylinder…) inside the workspace components images of the singularity-free cube. The idea will then be to optimize the location of the initial point $Q_0(\rho_{10},\rho_{20},\rho_{30})$ such that the image of the maximal singularity-free cube in the workspace generates a regular volume of maximal size.

### VI. Conclusion

A procedure for computing joint space singularities of *3-RPR* parallel manipulators has been presented firstly in this paper. Secondly, a procedure for the determination of maximal joint space singularity-free boxes has been provided.

These two procedures are of interest for planning trajectories in the joint space, and for manipulators design because they provide a tool for choosing the values of the joint limits.

Future work will optimize the choice of the cube center point $Q_0$ in the joint space in order to maximize the volumes of the workspace components images of the singularity-free cube.